\documentclass{article}[11pt] 

\usepackage{listings}
  
\usepackage[utf8]{inputenc} % allow utf-8 input
\usepackage[T1]{fontenc}    % use 8-bit T1 fonts
\usepackage{hyperref}       % hyperlinks
\usepackage{url}            % simple URL typesetting
\usepackage{graphicx}
\usepackage{amssymb}  

\title{A Novel Regularization Approach to Fair ML}
\author{
  Norman Matloff \\
  Department of Computer Science\\
  University of California, Davis\footnote{NM is professor of computer
  science, UCD, and has served as an expert witness in various
  litigation cases involving discrimination.} \\
  \texttt{matloff@cs.ucdavis.edu} \\
  and \\
  Wenxi Zhang \\
  University of California, Davis\footnote{WZ performed this work as an
  undergraduate at UCD.  She is now a graduate student in Data
  Science at Columbia University.} \\
  {\tt zwenxi22@gmail.com}
}

\begin{document} 

\maketitle

\begin{abstract}

A number of methods have been introduced for the fair ML issue,
most of them complex and many of them very specific to the underlying ML
moethodology.  Here we introduce a new approach that is simple, easily
explained, and potentially applicable to a number of standard ML
algorithms.  Explicitly Deweighted Features (EDF) reduces the impact of
each feature among the proxies of sensitive variables, allowing a
different amount of deweighting applied to each such feature.  The user
specifies the deweighting hyperparameters, to achieve a given point in
the Utility/Fairness tradeoff spectrum.  We also introduce a new,
simple criterion for evaluating the degree of protection afforded by any
fair ML method.

\end{abstract}

\section{Introduction}

There has been increasing concern that use of powerful ML tools may act
in ways discriminatory toward various protected groups, a concern that
has spawned an active research field.  One challenge has been that the
term ``discriminatory'' can have a number of different statistical
meanings, which themselves can be subject of debate concerning the
propriety of applying such criteria.

A common criterion in the litigation realm is that two individuals who
are of different races, genders, ages and so on but who are otherwise
similarly situated should be treated equally \cite{shrm}. As the Court
noted in \textit{Carson v.\ Bethlehem Steel }\cite{beth},

\begin{quote}  

The central question in any employment-discrimination case is whether
the employer would have taken the same action had the employee been of a
different race (age, sex, religion, national origin, etc.) and
everything else had remained the same

\end{quote}

Many US regulatory agencies apply a broader standard, the \textit{4/5
Rule} \cite{shrm}, and as noted, various criteria have been proposed in
the ML fairness literature.  However, the present paper focuses on the
above ``if everything else had remained the same'' criterion of
\textit{Carson v.\ Bethlehem Steel}, which is also the context of
\cite{lum} and \cite{scutari}.  This is the notion of \textit{disparate
treatment} \cite{shrm}.  See further discussion of legal aspects in
\cite{gray}, where the author is both a statistician and a lawyer. 

Disparate treatment can be described statistically as follows.  Consider
data of the form $(Y,X,S)$, for an outcome variable $Y$, a feature set
$X$ and a separate set of sensitive/protected variables $S$.  Say we
have a new case, $(Y_{new},X_{new},S_{new})$, with $Y_{new}$ to be
predicted.  Let $\widehat{Y}_{new}$ denote our predicted value.  Then
ideally, $\widehat{Y}_{new}$ will be statistically independent of
$S_{new}$ \cite{lum}.

However, note our use of the word \textit{ideally} above.  Full
independence of $\widehat{Y}$ and $S$ is achievable only under very
restrictive assumptions, as in \cite{lum} and \cite{scutari}.  In
particular, an obstacle exists in the form of a set of features $C$
belonging to $X$ that are correlated with $S$, resulting in indirect
information about $S$:  

On the other hand, the features in $C$ may have substantial predictive
power for $Y$.  One would not wish to exclude a feature that is strongly
related to $Y$ simply because it is somewhat related to $S$.  In other
words, we have the following issue. 

\begin{quote}
\textbf{The Fairness-Utility Tradeoff
\cite{zhao} \cite{moritz}:}
There is an inherent tradeoff betweent fairness and utility 
\end{quote}

Thus the set $D$ is the crux of the ML fairness issue.  How do we use
$C$ and the rest of $X$ to predict $Y$ well (utility), while using only
minimal information about $S$ (fairness)?

Though the fairness literature is already quite extensive, much of it is
either highly complex, or closely tied to specific ML methods, or both.
Our goals in the present work are to develop fair methods that

\begin{itemize}

\item are simple to explain and use,

\item apply to a number of standard ML algorithms, and 

\item do not require use of, or even knowledge of $S$.\footnote{That
last point refers to the fact that if we believe the features in $C$ are
related to $S$, we do not actually need data on $S$ itself.} 

\end{itemize} 

We present a general approach.  Explicitly Deweighted Features (EDF)
reduces the impact of $C$.  The term \textit{explicit} alludes to the
fact that the user drectly specifies the deweighting hyperparameters
$\delta_i$, one for each feature in $C$.

Our work is inspired by \cite{komiyama} and \cite{scutari}, which deal
with linear models.  Here EDF borrows from the well-understood
methodologies of ridge regression and the LASSO.  Moreover, we also will
present our analyses of random forests and the k-Nearest Neighbor
methods, in which EDF will deweight features in other manners natural
to the structure of the ML methods.

We will also introduce a novel measure of the strength of relation
between $\widehat{Y}$ and $S$ (the weaker, the better).  Recalling our
ideal criterion above of independence between these two quantities,
one might measure their correlation.  However, correlation is a less
desirable measure in cases in which one or both of the quantities is a
one-hot (indicator) variable.  Instead, we correlate modified versions
of the quantities, to be presented below.  Note that the modified
$\widehat{Y}$ will be based only on $C$ and the rest of $X$, not $S$,
thus forming a measure of fairneess dealing purely with $C$.

The organization of this paper is as follows.  We set the stage in
Section \ref{notation} regarding notation and statistical issues.  A
literature review follows next, in Section \ref{prev}, and we describe
our novel fairness measure in Section \ref{eval}. 
The linear/generalized linear portion of the paper then begins in
Section \ref{edflin}, with empirical evaluations in Section
\ref{empirlin}.  The material on other ML algorithms then starts in
Section \ref{otherml}.

% Indeed, in many cases we may have no training data on $S$ at all.
% However, it still would be desirable to develop a prediction rule that
% does not create disparate impact.

\section{Notation and Context}
\label{notation}

Except when otherwise stated, all mathematical symbols will refer to
population quantities, rather than sample estimates.\footnote{This
population/sample paradigm is of course the classical statistical view.
Readers who prefer to use ML terms such as \textit{probability
generative mechanism} may substitute accordingly.}

We consider here the common case in which $\widehat{Y}$ is derived from
an estimated regression function.  By ``regression function'' we simply
mean conditional expectation.  Notationally, it is convenient to use the
random variable form:  In $W = E(V|U)$, $U$ is considered random, and
thus so is $W$\cite{ross}.

The regression function is estimated from training data either via a
parametric (e.g.\ linear) or nonparametric (e.g.\ random forests)
approach.  And we of course include the binary/categorical case, in
which $Y$ is a dummy/one-hot indicator variable or a vector of such
variables, conditional probabilities.

In other words, ideally we aim to have $\widehat{Y}_{new} = E(Y_{new} |
X_{new})$ independent of $S_{new}$, or at least approximately so.  (In
the case of dichotomous $Y$, the analyst will typically do something
like rounding $\widehat{Y}$ to 0 or 1.) To avoid equation clutter, let
us drop the \textit{new} subscript and simply write the ideal condition
as 

\begin{equation}
\label{yhaindepts}
E(Y ~|~ X) \textrm{ and } S 
\textrm{ are statistically independent}
\end{equation}

We assume here that policy is to completely exclude $S$ from the ML
computation.  This is a slightly stronger assumption than those
typically made in the fairness literature, and we make the assumption
first for convenience; it simplifies notation and code.  Our methodology
would need only slight change below if $S$ were included, so this
assumption should not be considered a restriction.

And, second, the assumption makes sense from a policy point of view.
The analyst may, for instance, be legally required to exclude data on
$S$ during the training stage, i.e.\ to develop prediction rules based
only on estimation of $E(Y | X)$ rather than of $E(Y |
X,S)$.\footnote{Presumably the analyst \textit{is} allowed to use the
$S$ data in the training set to gauge how well her resulting prediction
rule works and test data.}

\section{Previous Work on Regularized Approaches to Fair ML}
\label{prev}

The term \textit{regularization} is commonly meant as minimization of a
loss function subject to some penalty, often in the form of $L_1$ and
$L_2$ penalties in linear regression models \cite{hastie}.  Our EDF
method falls into this category in the case of linear models,.

% https://arxiv.org/pdf/1707.00044v2.pdf
% https://arxiv.org/pdf/1806.06055.pdf                                            
% https://arxiv.org/pdf/1802.08626.pdf`    

In \cite{bechavod} the goal is to satisfy an \textit{equalized odds
criterion}. matching the rates of both false negatives and false
positives for all groups.  This is realized by adding terms to the
loss function that penalize the difference in the FPR and FNR between
the groups in the population.  Let us first review some of the
literature in this regard.

\cite{celis} proposes a meta-algorithm that incorporates a large set of
linear fairness constraints.  It runs a polynomial time algorithm that
computes an approximately optimal classifier under the fairness
constraints.  This approach is based on empirical risk minimization,
incorporating a fairness constraint that minimizes the difference of
conditional risks between two groups. Since this minimization is a
difficult nonconvex nonsmooth problem, they replace the hard loss in the
risk with a convex loss function (hinge loss) and the hard loss in the
constraint with the linear loss.

\cite{kamishima} proposed a (double-) regularization apprach based on
mutual information between Y and S, with special attention to indirect
prejudice, as is the case for the present paper.

\subsection{Relation to the Work of Komiyama \textit{et al} and Scutari
\textit{et al}}
\label{scutarietal}

A much-cited work that shares some similarity with that of the present
paper is that of \cite{komiyama}, later extended by for instance
\cite{scutari}.

In \cite{komiyama}, the usual least-squares estimate is modified by
imposing an upper bound on the coefficient of determination $R^2$
in the prediction of $Y$ by (a scalar) $S$.  Closed-form expressions are
derived, and a kernelized form is considered as well.  In this paper, we
focus on the approach of \cite{scutari}, which builds on that of
\cite{komiyama}.  

Those two papers develop a kind of 2-stage least squares method, which
we now describe in terms relevant to the present work.  We use different
notation, make previously-implicit assumptions explicit, and apply a bit
more rigor, but the method is the same.  As noted, all quantities here
are population values, not sample statistics.

The basic assumption (BA) amounts to $(Y,X,S)$ having a multivariate
Gaussian distribution, with $Y$ scalar and $X$ being a vector of length
$p$.  For convenience, assume here that $S$ is scalar.  All variables
are assumed centered.  The essence of the algorithm is as follows.

One first applies a linear model in regressing $X$ on $S$, 

\begin{equation}
E(X | S) = S \gamma 
\end{equation}

where $\gamma$ is a length-$p$ coefficient vector.  
BA implies the linearity.  

This yields residuals 

\begin{equation}
\label{resids}
U = X - S \gamma 
\end{equation}

Note that $U$ is a vector of length $p$.

$Y$ is then regressed on $S$ and $U$, which again by BA is linear:

\begin{equation}
\label{ysu}
E(Y ~|~ S,U) = \alpha S + \beta' U
\end{equation}

where $\alpha$ is a scalar and $\beta$ is a vector of length $p$.

Residuals and regressors are uncorrelated, i.e.\ have 0 covariance (in
general, not just in linear or multivariate normal settings).  So, from
(\ref{resids}), we have that $U$ is uncorrelated with $S$. In general 0
correlation does not imply independence.  but again under BA,
uncorrelatedness does imply independence.  This yields

\begin{equation}
E(S ~|~ U) = ES = 0
\end{equation}

and, by the Tower Property \cite{tower}, 

\begin{equation}
E(Y ~|~ U) = 
E \left [ E(Y ~|~ S,U) ~|~ U \right ] =
\beta' U
\end{equation} 

In other words, we have enabled $S$-free predictions of $Y$ by $U$,
i.e.\ (\ref{yhaindepts}).  The residuals $U$ form our new features,
replacing $X$. 

This technically solves the fairness problem, since $S$ and $U$ are
independent.  However, there are some major concerns:
\begin{itemize}

\item [(a)] The BA holds only rarely if ever in practical use.

\item [(b)] The above methodology, while achieving fairness, does so at
a significant cost in terms of our ability to predict $Y$.  This
methodology, in removing all vestige of correlation to $S$, may weaken some
variables in $X$ that are helpful in predicting $Y$.  In other words,
this methodology may go too far toward the Fairness end of the
Fairness-Utility Tradeoff.

\end{itemize} 

The authors in \cite{komiyama} were most concerned about (b) above.
They thus modified their formulation of the problem to allow some
correlation between $S$ and $\widehat{Y}$, the predicted value of $Y$.
They then formulate and solve an optimization problem that finds the
best predictor, subject to a user-specified bound on the coefficient of
determination, $R^2$, in predicting $S$ from $X$.  The smaller the
bound, the fairer the predictions, but the weaker the predictive
ability.  Their method is thus an approach to the Fairness/Utility
Tradeoff.

The later paper \cite{scutari} extends this approach, using a ridge
penalty.  This has the advantage of simplicity.  Specifically, they
choose $\widehat{\alpha}$ and $\widehat{\beta}$ to be

\begin{equation}
\textrm{argmin}_{a,b} ~ ||\mathbb{Y} - S a - \widehat{U} b||_{2}^2 +
\lambda ||a||_{2}^2
\end{equation}

where $\lambda$ is a regularizing hyperparameter.

If this produces a coefficient of determination at or below the desired
level, these are the final estimated regression coefficients.
Otherwise, an update formula is applied.

Thus \cite{scutari} finds a simpler, nearly-closed form extension of
\cite{komiyama}.  And they find in an empirical study that their method
has superior performance in the Fairness-Utility Tradeoff:  

\begin{quote}

We compare our approach with the regression model from Komiyama et al.
(2018), which implements a provably-optimal linear regression model,
and with the fair models from Zafar et al. (2019).
We evaluate these approaches empirically on six different data sets, and
we find that our proposal provides better goodness of fit and better
predictive accuracy for the same level of fairness. 

\end{quote}

Note carefully the difference between EDF and \cite{scutari} (and
\cite{komiyama}): 

\begin{itemize}

\item EDF allows limited but explicit involvement of the specific
features in $C$ in predicting $Y$. 

\item EDF completely excludes $S$, thus complying with European Union
regulations \cite{amstat}.  \cite{scutari} and \cite{komiyama} go much
further, allowing limited involvement of $S$ (even in the prediction
stage), thus possibly out of compliance with legal requirements in some
jurisdictions. 

\end{itemize} 

\section{Evaluation Criteria}
\label{eval}

As noted, the Fairness-Utility Tradeoff is a common theme in ML fairness
literature \cite{zhao} \cite{moritz}.  This raises the
question of how to quantify how well we do in that tradeoff.  In
particular, how do we evaluate fairness?  

Again, the literature is full of various measures for assessing
fairness, but here we propose another that some will find useful.
In light of our ideal state, (\ref{yhaindepts}), one might take a
preliminary measure to be 

\begin{equation}
\rho(\widehat{Y},S) = \textrm{ correlation between predicted Y and S}
\end{equation}

However, a correlation coefficient is generally used to measure
association between two continuous variables.  Yet in our context here,
one or both of $\widehat{Y}$ and $S$ may be one-hot variables, not continuous.
What can be done?  We propose the following.

\subsection{A Novel Fairness Measure}

As noted earlier, we aim for simplicity, with the measure of fairness
consisting of a single number if possible.  This makes the measure
easier to understand by nontechnical consumers of the analysis, say a
jury in litigation, and is more amenable to graphical display.  Our goal
here is thus to retain the correlation concept (as in \cite{scutari}),
but in a modified setting.

Specifically, we will use the correlation between modified versions of
$\widehat{Y}$ and $S$, denoted by $T$ and $W$.  They are defined as
follows:

\begin{itemize}

\item If $Y$ is continuous, set $T = \widehat{Y}$.  If instead $Y$ is
a one-hot variable, set $T = P(Y = 1 ~|~ X)$.

\item If $S$ is continuous, set $W = S$.  If instead $S$ is a one-hot
variable, set $W = P(S = 1 ~|~ X)$.

\end{itemize} 

Our fairness measure is then constructed by calculating
$\rho^2(T,W)$ for each component of $S$.

In other words, we have arranged things so that
in all cases we are correlating two continuous quantities. 
One then reports $\rho^2(T,W)$, estimated from our dataset.  The
reason for squaring is that squared correlation is well known to be the
proportion of variance of one variable explained by another (as in the
use of the coefficient of determination in \cite{komiyama} and
\cite{scutari}).

For categorical $S$, a squared correlation is reported for each
category, e.g.\ for each race if $S$ represents race.  

We assume that the ML method being used provides us with the estimated
value of $E(Y ~|~ X)$ in the cases of both numeric and
categorical $Y$; in the latter case, this is a vector of probabilities,
one for each category.  The estimated means and probabilities come
naturally from ML methods such as linear/generalized linear models,
tree-based models, and neural networks.  For others, e.g.\ support
vector machines, one can use probability calibration techniques
\cite{huang}.

%%% In all experiments, we apply cross-validation, with holdout size of 1000
%%% or 10\% of the data, whichever is smaller.  100 replications are
%%% performed.

\section{EDF in the Linear Case}
\label{edflin}

In the linear setting, we employ a regularizer $\delta$ that works as a
modified ridge regression operator, applied only to the variables in
$C$.  Note that here  $\delta$ will be a vector, allowing for different
amounts of deweighting for different features in $C$.

For $n$ data points in our training set, let $\mathbb{Y}$ denote the
vector of their $Y$ values, and define $\mathbb{X}$ similarly for the
matrix of $X$ values.   

\subsection{Structural Details}

Our only real asumption is linearity, i.e.\ that

\begin{equation}
E(Y | X) = \beta'X  
\end{equation}

for some constant vector $\beta$ of length $p$.  The predictors in $X$
can be continuous, discrete, categorical etc.  Since we are not
performing statistical inference, assumptions such as homoscedasticity
are not needed.

We then generalize the classic ridge regression problem to finding

\begin{equation}
\label{lindef}
\textrm{argmin}_b ~ 
||\mathbb{Y} - \mathbb{X} b||^2 + ||{D} b||^2
\end{equation}

where the diagonal matrix ${D} = diag(d_1,...,d_p)$ is a hyperparameter.
We will want $d_i = 0$ if the $i^{th}$ feature is not in $C$, while
within $C$, features that we wish to deweight more heavily will have
larger $d_i$.  (In order for $D$ to be invertible, we will temporarily
assume $d_i$ is positive but small for elements of $X$ not in $C$.)

Now with the change of variables $\widetilde{b} = {D} b$
and $\widetilde{\mathbb{X}} = \mathbb{X} {D}^{-1} $,
(\ref{lindef}) becomes

\begin{equation}
\label{lindef1}
\textrm{argmin}_{\widetilde{b}} ~ 
||\mathbb{Y} - \widetilde{\mathbb{X}} \widetilde{b}||^2 + 
||\widetilde{b}||^2 =
\textrm{argmin}_{\widetilde{b}} ~ 
||\mathbb{Y} - \widetilde{\mathbb{X}} \widetilde{b}||^2 + 
\lambda ||\widetilde{b}||^2 
\end{equation}

with $\lambda = 1$.  

Equation (\ref{lindef1}) is then in the standard ridge regression form,
and has the closed-form solution 

\begin{equation}
\widetilde{b} =
(\widetilde{\mathbb{X}}' \widetilde{\mathbb{X}} + \mathbb{I})^{-1}
\widetilde{\mathbb{X}}' \mathbb{Y}
\end{equation}

Mapping back to $b$ and $\mathbb{X}$, we have

\begin{equation}
\label{longmess}
b = {D}^{-1} \widetilde{b} = 
{{D}}^{-1}
 {[{{D}}^{-1} \mathbb{X}' \mathbb{X} {{D}}^{-1} + \mathbb{I}]}^{-1}
 {{D}}^{-1} \mathbb{X}' \mathbb{Y}
\end{equation}

Substituting $\mathbb{I} = {D} {{D}}^{-1}$ in (\ref{longmess}), and
using the identity $T^{-1} S^{-1} = (ST)^{-1}$, the above simplifies to

\begin{equation}
b = 
 {[{{D}}^{-1} \mathbb{X}' \mathbb{X} + {D}]}^{-1}
 {{D}}^{-1} \mathbb{X}' \mathbb{Y}
\end{equation}

and then, with $D = {{D}}^{-1} {DD} $, we have 

\begin{equation}
\label{d2}
b = 
 {[\mathbb{X}' \mathbb{X} + {D}^2]}^{-1}
 \mathbb{X}' \mathbb{Y}
\end{equation}

We originally assumed $d_i >0$, but due to the fact that (\ref{d2}) is
continuous in the $d_i$, we can take the limit as $d_i \rightarrow 0$,
showing that 0 values are valid.  We now drop that assumption.

In other words, we can achieve our goal---deweighting only in $C$, and
in fact with different weightings within $C$ if we wish---via a simple
generalization of ridge regression.\footnote{In a different context, see
also \cite{wu}.} We can assign a different $d_i$ value to each feature
in $C$, according to our assessment (whether it be \textit{ad hoc},
formal correlation etc.) of its impact on $S$.  We then set $d_i = 0$
for features not in $C$.  

Note again that the diagonal matrix $D$ is a hyperparameter, a set of
$|C|$ quantities $d_i$ controlling the relative weights we wish to place
within $C$.  We then set our deweighting parameters $\delta_i = d_{i}^2$.

Applying the well-known dual problem in this context, minimizing
(\ref{lindef}), the above is equivalent to choosing $b$ as

\begin{equation}
\label{ridge}
||\mathbb{Y} - \mathbb{X}b||^2 ~ \textrm{ subject to }
\sum_i (d_i b_i)^2 \leq \gamma
\end{equation}

where $\gamma$ is also a hyperparameter. 

\subsection{Computation}
\label{computlin}

One computational trick in ridge regression is to add certain artificial
data points, as follows.

Add $p$ rows to the feature design matrix, which in our case means

\begin{equation}
\label{eqna}
A = 
\left (
\begin{array}{cc}
\mathbb{X}  \\
D
\end{array}
\right )
\end{equation}

Add $p$ 0s to $\mathbb{Y}$:

\begin{equation}
\label{eqnb}
B = 
\left (
\begin{array}{cc}
\mathbb{Y}  \\
0
\end{array}
\right )
\end{equation}

Then compute the linear model coefficients as usual,

\begin{equation}
\label{ridgetrick}
(A'A)^{-1} A'B
\end{equation}

Since

\begin{equation}
A'A = 
\mathbb{X}'\mathbb{X} + D^2
\end{equation}

and 

\begin{equation}
A'B = \mathbb{X}'Y
\end{equation}

(\ref{ridgetrick}) gives us (\ref{d2}), our desired ridge estimator.  We
then can use ordinary linear model software on the new data (\ref{eqna})
and (\ref{eqnb}).

\subsection{Generalized Linear Case}
\label{lin}

In the case of the generalized linear model, here taken to be logistic,
the situation is more complicated.

Regularization in general linear models is often implemented by
penalizing the length of the coefficient vector in maximizing the
likelihood function \cite{glmnet}.  Also, \cite{kamishima} uses
conjugate gradient for solving for the coefficient vector.  In the EDF
setting, such approaches might be taken in conjunction with a penalty
term involving $||Db||^{2}$ as above.

As a simpler approach, one might  add artificial rows to the design
matrix, as in Section \ref{computlin}.  We are currently investigating
the efficacy of this \textit{ad hoc} approach.

\section{Issues Involving C}

\subsection{Choosing C}   

Which features should go into the $C$ deweighting set?  These are
features that predict $S$ well.  The analyst may rely on domain
expertise here, or use any of the plethora of feature engineering
methods that have been developed in the ML field.  In some of the
experiments reported later, we turned to FOCI, a model-free method for
choosing features to predict a specified outcome variable \cite{mona}.   

\subsection{Choosing the Deweighting Parameters}

The deweighting parameters can be chosen by cross-validation, finding
the best predictive ability for $Y$ for a given desired level for the
fairness measure $\rho^2$.  But if $C$ consists of more than one
variable, each with a different deweighting parameter, we must do a grid
search.  Our software standardizes the features, which puts the various
deweighting parameters on the same scale, and the analyst may thus
opt to use the same parameter value for each variable in $C$.  The
search then becomes one-dimensional.  Note that even in that case, EDF
differs from other shrinkage approaches to fair ML, since there is no
shrinkage associated with features outside $C$.

Our examples below are only exploratory, illustrating the use of EDF for
a few values of the deweighting parameters; no attempt is made to
optimize.

\section{Derivative Benefits from Use of a Ridge Variant}
\label{bonus}

It has long been known that ridge regression, although originally developed
to improve stability of estimated linear regression coefficients, 
also tends to bring improved prediction accuracy \cite{jf}.  Thus, while
in fair ML methods we are prepared to deal with a tradeoff between
accuracy and fairness, with EDF we may actually achieve improvements in
both aspects simultaneously.

\section{Empirical Investigation of Linear EDF}
\label{empirlin}

Note that software used here is \textbf{EDFfair} \cite{edf}.

\subsection{Census Dataset} 

Let us start with a dataset derived from the 2000 US census, consisting
of salaries for programmers and engineers in California's Silicon
Valley.\footnote{This is the \textbf{pef} data in the R
\textbf{regtools} package.}   Census data is often used to investigate
biases of various kinds.  Let's see if we can predict income without
gender bias.

Even though the dataset is restricted to programmers and engineers, it's
well known that there is a quite a variation between different job types
within that tech field.  One might suspect that the variation is
gendered, and indeed it is.  Here is a comparison between men and women
among the six occupations, showing the gender breakdown across
occupatioms:

\begin{tabular}{|r||r|r|r|r|r|r|}
\hline
gender & occ 100 & occ 101 & occ 102 & occ 106 & occ 140 & occ 141 \\ \hline
men & 0.2017 & 0.2203& 0.3434 & 0.0192& 0.0444& 0.1709 \\ \hline
women & 0.3117& 0.2349& 0.3274& 0.0426 & 0.0259 & 0.0575 \\ \hline
\hline
\end{tabular}

% \begin{lstlisting}[basicstyle=\small]
% men:
% 
%        100        101        102        106        140        141 
% 0.20168621 0.22032670 0.34336715 0.01923330 0.04446055 0.17092610 
% \end{lstlisting}
% 
% \begin{lstlisting}[basicstyle=\small]
% women:
% 
%        100        101        102        106        140        141 
% 0.31173594 0.23492258 0.32742461 0.04258354 0.02587612 0.05745721 
% \end{lstlisting}

Men and women do seem to work in substantially different
occupations.  A man was 3 times as likely as a woman to be in
occupation 141, for instance. 

Moreover, those occupations vary widely in mean wages:

\begin{tabular}{|r|r|r|r|r|r|}
\hline
occ 100 & occ 101 & occ 102 & occ 106 & occ 140 & occ 141 \\ \hline
50396.47 & 51373.53 & 68797.72 & 53639.86 & 67019.26 & 69494.44  \\ \hline
\hline
\end{tabular}

So, the occupation variable is a prime candidate for inclusion in $C$.

% \begin{lstlisting}[basicstyle=\small]
%      100      101      102      106      140      141 
% 50396.47 51373.53 68797.72 53639.86 67019.26 69494.44 
% \end{lstlisting}

Thus, $Y$ will be income, $S$ will be gender, and we'll take $C$ to be
occupation.  Here are the Mean Absolute Prediction Errors and
correlations between predicted $Y$ and gender, based on results of 500
runs, i.e.\ 500 random holdout sets:\footnote{The default holdout size
was used, 1000.  The dataset has 20090 rows.  
% Here and below, the results are accurate to approximately two or three decimal places.
}

\begin{tabular}{rrr}
\hline
$D_i^2$ & MAPE & $\rho^2$ \\
\hline
1.00 & 25631.21 & 0.22 \\
4.00 & 25586.08 & 0.22 \\
25.00 & 25548.24 & 0.22 \\
625.00 & 25523.83 & 0.21 \\
5625.00 & 25576.15 & 0.18 \\
15625.00 & 25671.33 & 0.15 \\
\hline
\end{tabular}

Now, remember the goal here.  We know that occupation is related to
gender, and want to avoid using the latter, so we hope to use occupation
only ``somewhat'' in predicting income, in the spirit of the 
Fairness-Utility Tradeoff.  Did we achieve that goal here?

We see the anticipated effect for occupation:  Squared correlation
between $S$ and predicted $Y$ decreases as $D_i^2$ increases, with only
a slight deterioriation in predictive accuracy in the last settings, and
with predictive accuracy actually improving in the first few settings---as
anticipated in Section \ref{bonus}.  Further reductions in $\rho^2$
might be possible by expanding $C$ to include more variables.

%% We see above that yes, we can limit unfairness, in that we can limit the
%% correlation between gender and our predicted values.  What about the
%% utility aspect?  Let'see how much using occupation helps prediction in a
%% linear model:
%% 
%% \begin{lstlisting}
%% # use only occupation, not gender
%% > replicMeans(500,"qeLin(pef[,-4],'wageinc')$testAcc")  
%% [1]25556.47
%% # use only gender, not occupation
%% > replicMeans(500,"qeLin(pef[,-3],'wageinc')$testAcc")  
%% [1] 25834.77
%% # use neither gender nor occupation
%% > replicMeans(500,"qeLin(pef[-(3:4)],'wageinc')$testAcc")  
%% [1] 26046.05
%% 
%% \end{lstlisting}
%% 
%% We are not allowed to use gender, but adding occupation itself to our
%% feature set does indeed help.  

\subsection{Empirical Comparison to Scutari \textit{et al}}
\label{empscu}

%% The approach in \cite{scutari} has no analog of our covariates set $C$.
%% Instead, the effect of covariates is partially neutralized by the nature
%% of $S$ and $U$.  By contrast, for our method, one need not even have $S$
%% at the training stage; one just needs $X$, including $C$ (though without
%% data on $S$, one would not have an assessment of the efficacy of the
%% result).  And our method is not limited to the linear case.

The authors of \cite{scutari} have made available software for their
method (and others), in the \textbf{fairml} package.  Below are the
results of running their function \textbf{frrm()} on the census
data.\footnote{The package also contains implementations of the methods
of \cite{komiyama} and \cite{zafar}, but since \cite{scutari} found
their method to work better, we will not compare to the other two.}
The `unfairness' parameter, which takes on values in [0,1], corresponds
to their ridge restriction; smaller values place more stringent
restriction on $||\alpha||$ in (\ref{ysu}), i.e.\ are fairer.  We set
their $\lambda$ parameter to 0.2.

\begin{tabular}{rrr}
\hline
\textrm{unfairness} & MAPE & $\rho^2$ \\
\hline
0.35 & 25538 & 0.2200 \\ 
0.30 & 25483 & 0.2193 \\ 
0.25 & 25353 & 0.2194 \\ 
0.20 & 25322 & 0.2198 \\ 
0.15 & 25484 & 0.2203 \\ 
0.10 & 25378 & 0.2193 \\ 
0.05 & 25339 & 0.2140 \\ 
\end{tabular}

The mean absolute prediction errors here are somewhat smaller than those
of EDF.  But the squared correlations between predicted $Y$ and
probabilities of $S$ (here, the probabilities of being female, the
sensitive attribute) are larger than those of EDF. In fact, the
correlation seemed to be constant, and no value of the fairness
parameter succeeded in bringing the squared correlation below 0.20.
Further investigation is needed.

% As noted, by construction $S$ and $U$ are uncorrelated, and if we are in
% a Gaussian situation, that implies independence.  Thus the squared
% correlations reported above may seem odd, but it actually is to be
% expected, as follows.  
% 
% The issue is one of unconditonal versus conditional independence.  Both
% the predicted $Y$ and probabilities of $S$ are conditional on $X$, and
% arguably those are the relevant quantities.  
% 
% Consider the example of the Pima diabetes data in the main paper, in
% which the sensitive variable was Age.  Another variable, Number of
% Pregnancies, was in our set $C$ of covariates to $S$.  Then if we know,
% say, that a woman has had 4 pregnancies, there is a strong likelihood
% that she is above a certain age.  If we wish our policy on some issue to
% not involve age, that implies that we should not base our policy on
% whether a woman has had many pregnancies.
% 
% The computation of the residuals $U$ is intended to remove the influence
% of the number of pregnancies.

\section{Other ML Algorithms}
\label{otherml}

The key point, then, is that $\ell_2$ regularization places bounds on
the the estimated linear regression coefficients $b_i$ (of course this
is true also for $\ell_1$ and so on)---but now for us the bounds are
only on the $b_i$ corresponding to the variables in $C$.  The goal,
as noted, is to have a ``slider'' one can use in the tradeoff between
utility and fairness.  Smaller bounds move the slider more in the
direction of fairness, with larger ones focusing more on utillty.
That leads to
our general approach to fairness:

\begin{quote}

For any ML method, impose fairness by restricting the amount of
influence the variables in $C$ have on predicted $Y$.

\end{quote}

\subsection{Approaches to Deweighting}

This is easy to do for both linear and several other standard ML methods:

\begin{itemize}

\item Random forests:  In considering whether to split a tree node, set
the probability of choosing variables in $C$ to be lower than for the
other variables in $X$.  The R package \textbf{ranger} \cite{ranger}
sets variable weights in this manner (in general, of course, not
specific to our fair-ML context here).

\item k-nearest neighbors (k-NN):  In defining the distance metric, place
smaller weight on the coordinates corresponding to $C$, as
allowed in \cite{yancey} (again, not specific to the fair-ML context). 

\item Support vector machines:  Apply an $\ell_2$ constraint on the portion
of the vector $w$ of hyperplane coefficients corresponding to $C$.  This
could be implemented using, for instance, the weight-feature SVM method
of \cite{wtdsvm} (once again, a feature not originally motivated by
fairness).

\end{itemize} 

As noted, in each of the above cases, the deweighting of features has
traditionally been aimed at improving prediction accuracy.  Here we
repurpose the deweighting for improving fairness in ML.

% Granted, almost any ML fairness method will reduce the influence of some
% features.  But here we are explicitly doing so, hence the name
% Explicitly Deweighted Features. 

\subsection{Empirical Investigation}

\subsubsection{The COMPAS Dataset}

This dataset has played a major role in calling attention to the need
for fair ML.\footnote{The version used was that of the \textbf{fairml}
package.  This is a highly complex issue.  See \cite{kl} for the many
nuances, legal and methodological.} The outcome variable $Y$ is an
indicator for recidivism within two years.

As with other studies, $S$ will be race, but choice
of $C$ is more challenging in this more complex dataset.

In predicting the race African-American, of special interest in the
COMPAS controversy, the relevant variables were found by FOCI to be, in
order of importance,

\begin{lstlisting}
 decile_score
   sex.Female
 priors_count
          age
  out_custody
juv_fel_count
     sex.Male
\end{lstlisting}

The \textbf{decile\_score} is of particular interest, as it has been at
the COMPAS controversy \cite{kl}.  There are indications that it may be
racially biased, which on the one hand suggests using it in $C$ but on
the other hand suggests using a small deweighting parameter.  In this
example, we chose to use decile\_score, gender, priors\_count and age.  

Here we used a random forests analysis.  We used a common deweighting
factor (1.0 means no reduction in importance of the variable) on these
variables, and default values of 500 and 10 for the number of trees and
minimum node size, respectively.

Below are the results, showing the overall probability of misclassification.
There were 100 replications for each line in the table. 

\begin{tabular}{rrr}
\hline
deweight factor & OPM & $\rho^2$ \\ 
\hline
1.00 & 0.2150 & 0.3136 \\ 
0.80 & 0.2109 & 0.3029 \\
0.60 & 0.2118 & 0.3028 \\ 
0.40 & 0.2116 & 0.2870 \\ 
0.20 & 0.2126 & 0.2770 \\ 
0.10 & 0.2187 & 0.2488 \\ 
0.05 & 0.2126 & 0.2484 \\ 
0.01 & 0.2203 & 0.2210 \\ 
\end{tabular}

Deweighting did reduce correlation between predicted $Y$ and $S$, with
essentially no compromise in predictive ability.  Interestingly, there
seems to be something of a ``phase change,'' i.e.\ a sharp drop,between
$C$ values 0.20 and 0.10.

The $;\rho^2$ values are still somewhat high, and expansion of $C$
may be helpful.

\subsubsection{The Mortgage Denial Dataset}

This dataset, from the \textbf{Sorted Effects} package, consists of data
on mortgage applications in Boston in 1990.  We predict whether the
mortage application is denied based on a list of credit-related
variables.  Here we take the indicator for Black applicant to be the
sensitive variable. The $Y$ variable here is an indicator variables for
mortgage application denial.  

Again choosing C is challenging in this complex dataset, so we run FOCI
to find relevant variables by predicting the race Black. The chosen
relevant varibles in the order of importance are 

\begin{lstlisting}
loan_val			
mcred			
condo
\end{lstlisting}

In this case, after experiments, we chose to just use loan\_val as the
variable $C$.  

Here we used k-nearest neighbors (k-NN).  The default value of 25 was
used for \textbf{k}.  Again we use the deweighting factor for variable
$C$ (1.0 means no reduction in importance of the variable).

\begin{tabular}{rrr}
\hline
deweight factor & OPM & $\rho^2$ \\
\hline
1.00 & 0.0953 & 0.0597 \\
0.80 & 0.0977 & 0.0621 \\
0.60 & 0.1012 & 0.0597 \\
0.40 & 0.1006 & 0.0563 \\
0.20 & 0.1189 & 0.0305 \\
0.10 & 0.2461 & 0.0373 \\
0.05 & 0.2820 & 0.0267 \\
0.01 & 0.3249 & 0.0238 \\
\end{tabular}

% \begin{tabular}{|r|r|r|}
%   \hline
%   deweighting factor & OPM & $\rho^2$ \\ \hline 
%   \hline
%   1.0 &0.1011 & 0.0622 \\ \hline
%   0.8 & 0.1141 & 0.0541 \\ \hline
%   0.6 & 0.1151 & 0.0519 \\ \hline
%   0.2 & 0.1151 & 0.0470 \\ \hline
%   \end{tabular}

We can see that the deweighting factor reduces the correlation between
$S$ and predicted $Y$, with little impact on predictive accuracy up
through a deweighting factor of 0.20.  Again, there appear to be phase
changes in both columns.

Next we also compute the correlation between fitted values of full
model(with s) and the EDF kNN model with deweighting variable C:

\begin{tabular}{rrr}
\hline
deweight factor & black.0 $\rho^2$ & black.1 $\rho^2$ \\
\hline
1.0 & 0.7865 & 0.7751 \\
0.8 & 0.7824 & 0.7756 \\
0.6 & 0.7754 & 0.7681 \\
0.2 & 0.7363 & 0.7793 \\
\end{tabular}
    
Here we can see that overall the correlations for all levels (black.0 and
black.1) are high.  As we deweight more, the correlation decrease a
little. Thus we can say that Y isn't strongly related to the sensitive
variable Black and C have relative substantial predictive power for Y.
Thus we can see that loan\_val is an appropriate proxy for the sensitive
variable Black.

We also apply random forests analysis to Mortgage dataset, with C variable being loan\_val. Below are the results:

\begin{tabular}{rrr}
  \hline
  deweight factor & OPM & $\rho^2$ \\ 
  \hline
  1.0 & 0.0943 & 0.06953 \\ 
  0.8 & 0.0882 & 0.0602 \\ 
  0.6 & 0.0910 & 0.0523 \\ 
  0.3 & 0.0924 & 0.0434 \\ 
  \end{tabular}

We have similar results for random forests analysis. We see a decrease
in correlation as we deweight the $C$ variable.  We even has a higher overall
accuracy in the random forests analysis. For future analysis, we may
also apply deweighting on different ML methods to see which method best
suit the dataset.  And again, using more features in $C$ would likely be
helpful.

\section{Conclusions and Future Work}
\label{future}

In the examples here, and others not included here, we have found that EDF
can be an effective tool for analysts hoping to achieve a desired point
in the Utility/Fairness tradeoff spectrum.   More types of data need to
be explored, especially those in which the set $C$ is large.

Ridge regression, feature weighting in random forests and k-NN, and so
on, were orginally developed for the purpose of improved prediction and
estimation.  We have repurposed them here in a novel setting, seeking
fair ML. 

We have also introduced a new approach to evaluating fairness, involving
correlations of probabilities.  It will be interesting to see how the
various fair ML methods fare under this criterion. 

As noted, R software for our methods is publicly available.

\section{Author Contributions}

NM conceived of the problem, derived the mathematical expressions, and
wrote the initial version of the software.  WZ joined the project
midway, but made significant contributions to the experimental design,
software and analysis of relations to previous literature.

\begin{table}[!htbp] \centering 
  \caption{} 
  \label{} 
\begin{tabular}{@{\extracolsep{5pt}} cc} 
\\[-1.8ex]\hline 
\hline \\[-1.8ex] 
sex & MAPE \\ 
\hline \\[-1.8ex] 
$0.111$ & $25,491.260$ \\ 
$0.111$ & $25,816.110$ \\ 
$0.114$ & $25,693.020$ \\ 
$0.106$ & $25,708.600$ \\ 
$0.111$ & $25,789.700$ \\ 
$0.110$ & $25,959.280$ \\ 
$0.114$ & $25,931.790$ \\ 
$0.115$ & $25,856.950$ \\ 
\hline \\[-1.8ex] 
\end{tabular} 
\end{table}

\bibliographystyle{acm}
\bibliography{Current}

\end{document}